\documentclass[runningheads]{llncs}
\usepackage{graphicx}
\begin{document}
\title{From Modelling to Understanding Children’s Behaviour in the Context of Robotics and Social Artificial Intelligence}
\author {Serge Thill\inst{1}\orcidID{0000-0003-1177-4119} \and
Vicky Charisi\inst{2}\orcidID{0000-0001-7677-027X} \and
Tony Belpaeme\inst{3}\orcidID{0000-0001-5207-7745} \and
Ana Paiva \inst{4}\orcidID{0000-0003-3998-5188}}
\authorrunning{Thill et al.}
\institute{Donders Institute for Brain, Cognition, and Behaviour, Radboud University, Nijmegen, the Netherlands \and
Joint Research Centre, European Commission, Seville, Spain \and
%\url{http://www.springer.com/gp/computer-science/lncs} 
Ghent University and imec, Ghen, Belgium\\
\and GAIPS, Instituto Superio T\'{e}cnico, University of Lisbon, Lisbon, Portugal
}
\maketitle              % typeset the header of the contribution
\begin{abstract}
%The abstract should briefly summarize the contents of the paper in 15--250 words.

Understanding and modelling children's cognitive processes and their behaviour in the context of their interaction with robots and social artificial intelligence systems is a fundamental prerequisite for meaningful and effective robot interventions. However, children's development involve complex faculties such as exploration, creativity and curiosity which are challenging to model. Also, often children express themselves in a playful way which is different from a typical adult behaviour. Different children also have different needs, and it remains a challenge in the current state of the art that those of neurodiverse children are under-addressed. With this workshop, we aim to promote a common ground among different disciplines such as developmental sciences, artificial intelligence and social robotics and discuss cutting-edge research in the area of user modelling and adaptive systems for children.

\end{abstract}

\section{Overview}
Robots and social artificial intelligence systems can support children's development and well-being by providing automatic, real-time and personalised feedback in the context of natural interactions ~\cite{irfan2019personalization}. These technological advances have the potential to positively transform various areas related to children's development, such as the formal education, of neurotypical and neurodiverse children. However, for a meaningful and successful personalised intervention, these systems need to develop an understanding of the user, a mental model ~\cite{tabrez2020survey} ~\cite{thill2012robot} that integrates the child's individual characteristics, their needs and preferences, their current reactions as well as their development over the course of a long term interaction. As such, modelling children's behaviour is a challenging endeavour particularly if we consider children's intrinsic motivation for exploration, play, creativity and curiosity, necessary faculties for their holistic development, that often make them distinct from adults.

Decades of research in developmental science have produced an increasingly detailed characterisation of learning in children and provide the fundamentals for our understanding of their behaviour and development. More recently, the use of computational approaches in developmental science provides new insights in children's behaviours and cognitive processes which sometime is in contrast with previously established beliefs about children's development. Recent research, for example, criticises the picture of human life history as involving a linear transition from more curious in early childhood to less curious with age. Instead, exploration appears to become more elaborate throughout human childhood ~\cite{pelz2020elaboration}. Furthermore, research with neurodiverse children has shed light to our understanding of atypical cognitive processes and social and emotional engagement, language learning and play ~\cite{pellicano2022annual} ~\cite{gargot2022automatic}.

At the same time, recent developments in artificial intelligence have resulted in breakthrough improvements in areas like  machine learning, computer vision and speech processing. These developments have made it increasingly easy to develop applications in robotics, too; for example, libraries such as OpenCV and OpenPose allow for simplified image processing and generation of data sets of child-robot interactions ~\cite{lemaignan2018pinsoro}.

At the same time, these new solutions raise new challenges, for example with respect to biases in data sets and algorithms \cite{mehrabi2021surveyhr} that mean any solution developed using those might not work for aspects that are not considered as much, such as specific needs of neurodiverse children. Attempts to quantify and assess personality traits such as trust are also very problematic ~\cite{spanton2022measuring} in this context, especially since they might lead to a misguided notion that automated assessment and classification of children might be possible. More generally, policy initiatives and guidelines indicate that in parallel with the emerging opportunities robotics and social AI bring for children, there are risks that need to be addressed during the whole life-cycle of an AI product ~\cite{unicef2021} ~\cite{JRC127564}.

%EU regulations consider AI solutions deployed for children to be a high risk activity that needs to underlie particular ethical and regulatory scrutiny.

Overall, while progress in AI and ML opens vast new opportunities for deploying truly novel solutions (robotic or not) that can have a positive impact on the lives of children, the road there is far from straightforward and there remains an urgent need to raise awareness of the challenges that lie along it as well as discussing how to overcome them. 

There also remain fundamental technical challenges; for example, it is still not clear how to design algorithmic (and possibly robotic) approaches towards Theory of Mind despite decades of attempts ~\cite{scassellati2002theory}. Here, research and developments in the field of social robotics and child-robot interaction also brings new insights but also new questions regarding personalisation in children's development. Robots have proven effective in cognitive and socio-emotional support for second language learning ~\cite{vogt2019second}, problem-solving ~\cite{charisi2020child} and story-telling \cite{park2017telling}. In addition, researchers are trying to understand the impact of robots on child-child social interaction ~\cite{charisi2021effects}, perspective taking ~\cite{yadollahi2022motivating} and inter-generational interaction ~\cite{joshi2019robots} as well as a tool to study flexibility in human social cognition ~\cite{wykowska2020social}. However, often current research focuses on very controlled tasks, and modelling children's processes in complex real-life environments remains still a challenge. 

As such, the main questions we aim to tackle with this workshop are the following: How can we leverage the insights from developmental science to create social artificial agents able to model children's complex behaviour for meaningful and effective interventions? How might these models help us understand even better children's behaviour especially in the context of their interaction with artificial agents? How can we ensure that resulting technological solutions are in line with children's needs in a fair and non-discriminatory way and mitigate any emerging risk related to AI? How can we take advantage by the involvement of relevant stakeholders such as educators and industry?

%How can we ensure that resulting technological solutions serve the needs of all children and what are the risks for abuse?

The overarching aim of this workshop is to gain interdisciplinary insights into these questions and to promote a common ground and a shared understanding among the relevant scientific communities through a program consisting of paper presentations, interactive sessions, invited talks and a panel discussion.

% \vspace*{-4mm}
\subsection{Activities, Schedule and Format}
% \vspace*{-3mm}
 
Submission of 4-6 pages papers describing preliminary results or work in progress relevant to the workshop is encouraged. Submitted papers will undergo a review process. We propose a full day workshop that will include two sessions of short paper presentations from accepted submissions (child's development, AI algorithms and robot solutions), oral presentations from the invited speakers, two interactive sessions and to stimulate interaction among participants, we will culminate the workshop with a plenary discussion and potentially social robot demonstrations.

We propose a full day workshop on the 16th of December with the following tentative schedule:

\textbf{Tentative Schedule\\}
9:00 Welcome by the organizers\\
9:15 Invited talk 1\\
10:00 Short papers presentations 1\\
10:20 Enlightening talk by educator 1 \\
10:30 Coffee break \\
10:45 Short papers presentations 2\\
11:15 Invited talk 2 \\
12:00 Interactive session 1 \\
12:30 Lunch break \\
13:30 Invited talk 3 \\
14:15 Enlighting talk by educator 2 \\
14:25 Enlighting talk by educator 3\\
14:35 Interactive session 2\\
15:15 Invited talk 4 \\
16:00 Plenary discussion with the participation of the audience\\
16:45 Wrap-up \\
17:00 End of the workshop \\

%\vspace*{-4mm}
\subsection{List of Possible Topics}
%\vspace*{-3mm}
The planned discussion topics within this workshop will include but will not be limited to the following:

\begin{itemize}
    \item User modelling
\item Artificial Intelligence in HRI
 \item Adaptive robots for children
 \item Evaluation methods founded on cognitive, developmental or comparative psychology
\item Child-robot interaction
\item Child speech recognition
\item Eye-tracking techniques
 \item Children's perspectives about robots
 \item Ethical considerations
 \item Privacy and data minimization

\end{itemize}

%\vspace*{-4mm}
\subsection{Target Audience and Pre-requisites}
%\vspace*{-3mm}
The workshop addresses a broad range of researchers within the fields of Social Robotics, HRI, HCI, Design, Developmental Robotics, that work with neurotypical and neurodiverse children neighboring disciplines. 

%\vspace*{-4mm}
\subsection{Recruitment and expected number of participants}
%\vspace*{-3mm}
Participants will be recruited by means of a call for papers distributed via relevant mailing lists, social media, the organizers professional network, the IEEE society for cognitive and developmental systems and via the International Consortium of Socially Intelligent Robotics (https://mypersonalrobots.org/). 
 
 %\vspace*{-4mm}
\subsection{Plan for Documenting the work}
 %\vspace*{-3mm}
 The accepted papers will be published as proceedings at the http://ceur-ws.org. A special issue in a suitable journal will be organised depending on the themes of the submissions received and the final outcomes of the workshop discussions.

\section{Invited Speakers}

Confirmed invited speakers include:

\begin{itemize}
    \item Thomas Weisswange (Principal Scientist, HONDA Research Institute EU)
     \item Mohamed Chetouani (Professor, Sorbonne Université, ISIR-UPMC, CNRS)
     \item Agnieszka Wykowska (Principal Investigator, Italian Institute of Technology)
\end{itemize}

In addition to the above mentioned invited speakers, we aim to stimulate a discussion by introducing three short talks by practitioners and educators, with the following confirmed speakers:

Enlighting talks by educators:

\begin{itemize}
    \item Tiija Rinta, University College London, UK
     \item Tomoko Imai, Jiyugaoka Gakuen High School, Tokyo, Japan
     \item Chris Zotos, Arsakeio Lyceum, Patras, Greece
\end{itemize}

%Hae Won Park? (Child modeling)

%Stephanos Nikolaidis? (User modeling and machine learning)

%Liz Pellicano? (autism)

%Takayuki Kanda? (child modeling)

\section{Organizers}

\begin{itemize}

\item Serge Thill is an associate professor of artificial intelligence at the Donders Institute, Radboud University. He is a cognitive roboticist and cognitive scientist with a background in computational modelling and computational neuroscience. He is interested in human cognition in the context of interaction with different artificial cognitive systems such as robots. His spans a range of disciplines, from theoretical cognitive science (in particular theories of embodiment and how these relate to machine intelligence) over language and concept grounding to (neuro)-computational models of cognitive mechanisms and practical applications in, for example, autonomous vehicles and robots for therapy for children with autism spectrum disorder. Most recently, he is a PI on the recently funded Horizon Europe Project EMPOWER, which develops technological support for neurodiverse children in educational settings.

\item Vicky Charisi (vasiliki.charisi@ec.europa.eu) \\ is Research Scientist at the Joint Research Centre of the European Commission with a focus on the impact of AI on human behaviour, with a particular interest on child-robot interaction. Her research includes topics such as child's cognitive and socio-emotional development in the context of their interaction with social robots. In parallel, her work informs policy-oriented discussions in relation to AI and children's rights and she is interested in bringing different stakeholders, including researchers, policymakers, children, developers etc together to create common understaning in temrs of AI and children's rights. She is an Associate Editor at the International Journal of Child-Computer Interaction and she serves as a Chair of the IEEE Computational Intelligence Society for Cognitive and Developmental Systems TF for Human-Robot Interaction.

\item Tony Belpaeme is Professor at Ghent University and Visiting Professor of Cognitive Systems and Robotics at Plymouth University. He is a member of IDLab – imec at Ghent and is associated with the Centre for Robotics and Neural Systems at Plymouth. His research interests include social systems, cognitive robotics,  and artificial intelligence in general.

\item Ana Paiva focuses on the problems and techniques for creating social agents that can simulate human-like behaviours, be transparent, natural and eventually, give the illusion of life. Over the years she has dealt with this problem by engineering agents that exhibit specific social capabilities, including aspects such as emotions, personality, culture, non-verbal behaviour, empathy, collaboration, and others. 

\end{itemize}

\section{Funding}

This workshop is supported in part by the Horizon Europe project EMPOWER (www.project-empower.eu), grant agreement No 101060918, funded by the European Commission.

\bibliographystyle{splncs04}
\bibliography{icsrsws}

\begin{thebibliography}{10}
\providecommand{\url}[1]{\texttt{#1}}
\providecommand{\urlprefix}{URL }
\providecommand{\doi}[1]{https://doi.org/#1}

\bibitem{JRC127564}
Charisi, V., Chaudron, S., Di~Gioia, R., Vuorikari, R., {Escobar Planas}, M.,
  {Sanchez Martin}, J., {Gomez Gutierrez}, E.: Artificial intelligence and the
  rights of the child: Towards an integrated agenda for research and policy.
  Scientific analysis or review KJ-NA-31048-EN-N (online), Luxembourg
  (Luxembourg) (2022). \doi{10.2760/012329 (online)},
  \url{https://publications.jrc.ec.europa.eu/repository/handle/JRC127564}

\bibitem{charisi2020child}
Charisi, V., Gomez, E., Mier, G., Merino, L., Gomez, R.: Child-robot
  collaborative problem-solving and the importance of child's voluntary
  interaction: a developmental perspective. Frontiers in Robotics and AI
  \textbf{7}, ~15 (2020)

\bibitem{charisi2021effects}
Charisi, V., Merino, L., Escobar, M., Caballero, F., Gomez, R., G{\'o}mez, E.:
  The effects of robot cognitive reliability and social positioning on
  child-robot team dynamics. In: 2021 IEEE International Conference on Robotics
  and Automation (ICRA). pp. 9439--9445. IEEE (2021)

\bibitem{unicef2021}
Dignum, V., Penagos, M., Pigmans, K., Vosloo, S.: Policy guidance on ai for
  children (draft). Tech. rep., {UNICEF} (2021),
  \url{https://www.unicef.org/globalinsight/media/2356/file/UNICEF-Global-Insight-policy-guidance-AI-children-2.0-2021.pdf.pdf}

\bibitem{gargot2022automatic}
Gargot, T., Archambault, D., Chetouani, M., Cohen, D., Johal, W., Anzalone,
  S.M.: Automatic assessment of motor impairments in autism spectrum disorders:
  a systematic review. Cognitive Computation pp. 1--36 (2022)

\bibitem{irfan2019personalization}
Irfan, B., Ramachandran, A., Spaulding, S., Glas, D.F., Leite, I., Koay, K.L.:
  Personalization in long-term human-robot interaction. In: 2019 14th ACM/IEEE
  International Conference on Human-Robot Interaction (HRI). pp. 685--686. IEEE
  (2019)

\bibitem{joshi2019robots}
Joshi, S., {\v{S}}abanovi{\'c}, S.: Robots for inter-generational interactions:
  implications for nonfamilial community settings. In: 2019 14th ACM/IEEE
  International Conference on Human-Robot Interaction (HRI). pp. 478--486. IEEE
  (2019)

\bibitem{lemaignan2018pinsoro}
Lemaignan, S., Edmunds, C.E., Senft, E., Belpaeme, T.: The pinsoro dataset:
  Supporting the data-driven study of child-child and child-robot social
  dynamics. PloS one  \textbf{13}(10),  e0205999 (2018)

\bibitem{park2017telling}
Park, H.W., Gelsomini, M., Lee, J.J., Breazeal, C.: Telling stories to robots:
  The effect of backchanneling on a child's storytelling. In: 2017 12th
  ACM/IEEE International Conference on Human-Robot Interaction (HRI. pp.
  100--108. IEEE (2017)

\bibitem{pellicano2022annual}
Pellicano, E., den Houting, J.: Annual research review: Shifting from ‘normal
  science’to neurodiversity in autism science. Journal of Child Psychology
  and Psychiatry  \textbf{63}(4),  381--396 (2022)

\bibitem{pelz2020elaboration}
Pelz, M., Kidd, C.: The elaboration of exploratory play. Philosophical
  Transactions of the Royal Society B  \textbf{375}(1803),  20190503 (2020)

\bibitem{scassellati2002theory}
Scassellati, B.: Theory of mind for a humanoid robot. Autonomous Robots
  \textbf{12}(1),  13--24 (2002)

\bibitem{spanton2022measuring}
Spanton, R.W., Guest, O.: Measuring trustworthiness or automating physiognomy?
  a comment on safra, chevallier, gr$\backslash$ezes, and baumard (2020). arXiv
  preprint arXiv:2202.08674  (2022)

\bibitem{tabrez2020survey}
Tabrez, A., Luebbers, M.B., Hayes, B.: A survey of mental modeling techniques
  in human--robot teaming. Current Robotics Reports  \textbf{1}(4),  259--267
  (2020)

\bibitem{thill2012robot}
Thill, S., Pop, C.A., Belpaeme, T., Ziemke, T., Vanderborght, B.:
  Robot-assisted therapy for autism spectrum disorders with (partially)
  autonomous control: Challenges and outlook. Paladyn  \textbf{3}(4),  209--217
  (2012)

\bibitem{vogt2019second}
Vogt, P., van~den Berghe, R., de~Haas, M., Hoffman, L., Kanero, J., Mamus, E.,
  Montanier, J.M., Oran{\c{c}}, C., Oudgenoeg-Paz, O., Garc{\'\i}a, D.H.,
  et~al.: Second language tutoring using social robots: a large-scale study.
  In: 2019 14th ACM/IEEE International Conference on Human-Robot Interaction
  (HRI). pp. 497--505. Ieee (2019)

\bibitem{wykowska2020social}
Wykowska, A.: Social robots to test flexibility of human social cognition.
  International Journal of Social Robotics  \textbf{12}(6),  1203--1211 (2020)

\bibitem{yadollahi2022motivating}
Yadollahi, E., Couto, M., Dillenbourg, P., Paiva, A.: Motivating children to
  practice perspective-taking through playing games with cozmo. In: 2022 31st
  IEEE International Conference on Robot and Human Interactive Communication
  (RO-MAN). pp. 1482--1489. IEEE (2022)

\end{thebibliography}

%Irfan, B., Ramachandran, A., Spaulding, S., Glas, D. F., Leite, I., and Koay, K. L. (2019, March). Personalization in long-term human-robot interaction. In 2019 14th ACM/IEEE International Conference on Human-Robot Interaction (HRI) (pp. 685-686). IEEE.

%\end{thebibliography}
\end{document}